\def\year{2021}\relax
\documentclass[letterpaper]{article} 
\usepackage{aaai21}  
\usepackage{times}  
\usepackage{helvet} 
\usepackage{courier}  
\usepackage[hyphens]{url}  
\usepackage{graphicx} 
\usepackage{natbib}  
\usepackage{caption} 
\usepackage{comment}
\usepackage{subcaption}
\usepackage{multirow}
\usepackage{booktabs}
\usepackage{subcaption}
\usepackage[table,xcdraw]{xcolor}

\title{Cross-Lingual and Cross-Domain Crisis Classification\\ for Low-Resource Scenarios}
\author {
    Cinthia Sánchez,\textsuperscript{\rm 1,2}
    Hernan Sarmiento,\textsuperscript{\rm 1,2}
    Andres Abeliuk,\textsuperscript{\rm 1,3}
    Jorge Pérez,\textsuperscript{\rm 1,2} 
    Barbara Poblete,\textsuperscript{\rm 1,2,3} \\
}
\affiliations {
    \textsuperscript{\rm 1}Department of Computer Science, University of Chile \\
    \textsuperscript{\rm 2}Millennium Institute for Foundational Research on Data (IMFD), Santiago, Chile \\
    \textsuperscript{\rm 3}National Center for Artificial Intelligence (CENIA), Santiago, Chile \\

    \textbraceleft cisanche, hsarmien, aabeliuk, jperez, bpoblete\textbraceright @dcc.uchile.cl
}

\begin{document}

\maketitle

\begin{abstract}
Social media data has emerged as a useful source of timely information about real-world crisis events.
One of the main tasks related to the use of social media for disaster management is the automatic identification of crisis-related messages.
Most of the studies on this topic have focused on the analysis of data for a particular type of event in a specific language.
This limits the possibility of generalizing existing approaches because models cannot be directly applied to new types of events or other languages.
In this work, we study the task of automatically classifying messages that are related to crisis events by leveraging cross-language and cross-domain labeled data. 
Our goal is to make use of labeled data from high-resource languages to classify messages from other (low-resource) languages and/or of new (previously unseen) types of crisis situations.
For our study we consolidated from the literature a large unified dataset containing multiple crisis events and languages.
Our empirical findings show that it is indeed possible to leverage data from crisis events in English to classify the same type of event in other languages, such as Spanish and Italian (80.0\% F1-score). 
Furthermore, we achieve good performance for the cross-domain task (80.0\% F1-score)
in a cross-lingual setting.
Overall, our work contributes to improving the data scarcity problem that is so important for multilingual crisis classification.
In particular, mitigating {\em cold-start} situations in emergency events, when time is of essence. 

\end{abstract}

\section{Introduction}
Social media has created new possibilities for outreach and communications with worldwide scope.
It enables users to share information and opinions rapidly to millions of others in just seconds.
The content published in these platforms is diverse in format (e.g., video, images, audio), but also diverse in terms of the users' backgrounds (e.g., language, geographic location, culture).
This one-to-many broadcast of information has played a key role in the dissemination of breaking news and, in recent years, of emergency events~\cite{palen2016crisis}.
Across the different phases of crisis management
, social media also plays a role in planning and training, collaborative problem solving and decision making, and information gathering \cite{rporole}.

The microblogging platform Twitter\footnote{\url{http://twitter.com}} has become one of the primary data sources for real-time crisis-event information.
Its short-text messages, called {\em tweets}, have been widely studied to extract useful and timely knowledge for crisis management \cite{CrisisLex6, CrisisLex26, SoSItalyT4, CrisisNlpR1, poblete2018robust, crisismmd}.
Accurate and well-timed information about a crisis allows emergency relief agencies to act quickly and effectively, thus reducing the negative impact on society \cite{Graf2018}.
In this context, automatically identifying user messages related to crises becomes relevant.
However, the noisy nature of social media and unreliable information quality are significant challenges in this field.
In particular, online conversations are short, imprecise, and cover a wide range of topics, with little to no contextual information \cite{Troudi2018}. 
Furthermore, text message processing can be complex from a Natural Language Processing (NLP) perspective because they contain informal language, abbreviations, spelling mistakes, multilingual constructions, among others \cite{Ghosh2018}. 
All of these issues are further exacerbated when considering multilingual content \cite{espol2019}.
%

As in many other supervised learning tasks, most research and models address English content, 
which affects the availability of resources across languages \cite{CinthiaDC, Khare2018, Benchmarking}.
%
%
In general, existing approaches require a significant amount of labeled data to learn effective models \cite{HumAID2021}.
Often, labeled data from previous disasters is available, but in languages or crisis domains different from the target event \cite{Li2020}. 
This is problematic when we address multilingual and cross-domain scenarios where labeled data is scarce, leading to {\em cold-start} problems in early stages of a disaster \cite{DBLP_Lorini, Li2020}. 
%
Furthermore, training individual models for all possible scenario combinations, in terms of languages and types of events, is computationally expensive~\cite{Khare2018}. 
Hence, it becomes critical to create ways to leverage existing knowledge about crisis situations to address new or diverse types of events. 

We address this problem by studying the task of classifying crisis messages across languages and types of crisis domains.
Specifically, we focus on the binary classification task of tweets that are {\em related} and {\em unrelated} to crises. 
%
We consider this particular task to be relevant because of the high volume of noisy data that is found in Twitter during crisis events.
This noise consists mostly of unrelated content, such as spam and advertising, that can be found even when message retrieval is done using event-specific keywords (such as hashtags or locations associated with the crisis) \cite{CrisisLex26, Graf2018}.
Identifying related and unrelated messages is one of the first and foremost tasks required by emergency practitioners when using social media data. 

We center our work in two main objectives: 1)~transferring knowledge from one or more crisis domains (e.g., {\em earthquakes}, {\em hurricanes}, etc.) to other crisis domains, and 2)~transferring knowledge from crises in one (high-resource) language to another (low-resource) language.
To achieve this, we perform a comprehensive evaluation of transfer learning performance, using different data representations and transfer learning scenarios. 
%

%
Our results show that it is possible to leverage crisis data from high-resource languages, such as English, to classify crisis events in other languages, like Spanish or Italian 
(80.0\% F1-score). 
In addition, we observe that information from one or more crisis domains can contribute to the classification of events from a new domain. 
These results indicate that some characteristics of crises cross-cut different domains and languages.
This will allow researchers and emergency practitioners to improve the reach of automated social network analysis during crises at a low cost.
\medskip

\noindent{\bf Contributions.} In summary, our contributions are: 
%
\begin{enumerate}
\item We introduce a new multilingual and multi-domain crisis dataset, containing 53 crisis events and more than 160,000 messages. This dataset is the result of our work to unify 7 publicly available crisis datasets, achieved by using a systematic approach to consistently merge labels, and to add specific event and message metadata.

%

\item We show empirically that it is possible to perform transfer learning, using crisis data from high-resource languages, to  classify data from other (low-resource) languages and from new crisis domains. This addresses the important {\em cold-start} problem in crisis classification for low-resource languages and new crisis domains. 

\item We perform a novel comparative study of 7 state-of-the-art data models for cross-domain and cross-language transfer learning scenarios.

\end{enumerate}

\smallskip
\noindent {\bf Note on reproducibility.}  All of the source code and data used in this article are publicly available at
\url{https://github.com/cinthiasanchez/Crisis-Classification}.

\medskip
\noindent {\bf Roadmap.} 
In Section \ref{sec:related_work}, we present the literature review focusing on crisis messages classification and transfer learning techniques. 
In Section \ref{sec:proposal}, we detail our proposed experimental framework. 
In Section \ref{sec:data_unification}, we explain the construction of the unified dataset.
In Section \ref{sec:setup_train_test}, we describe the experimental setup among different domains and languages, and in Section \ref{sec:results_language}, we report the results.
In Sections \ref{sec:discussion} and \ref{sec:conclusion}, we present the discussion and final comments of our work.

\section{Related Work} \label{sec:related_work}
In this section, we discuss relevant prior literature for our work.
%
In particular, we focus on two areas: \textit{classification of crisis-related messages} and \textit{transfer learning}.

\subsection{Classification of Crisis-Related Messages}
One of the most prevalent tasks in this area is the binary classification of messages that are {\em related} vs. those that are {\em not related} to a crisis~\cite{CrisisLex6, ChileEarthquakeT1, CrisisNlpR7, espol2019}.
In the literature, the term {\em related} to a crisis event is usually interchangeable with the terms {\em relevant} or {\em informative}. 
First, a message is considered as {\em related} to a crisis when it refers implicitly or explicitly to a specific disaster \cite{review_article}.
A message considered {\em relevant} to a crisis is one that contains actual information pertaining to the event \cite{review_article, ChileEarthquakeT1}. 
It also conveys or reports information useful for crisis response \cite{CrisisNlpR7}. 
Finally, a message is considered {\em informative} when it contributes to a better understanding of the situation on the ground \cite{CrisisLex26}.
These terms have certain differences, especially in their degree of generalization and usefulness for humanitarian aid.

\citet{CrisisLex6} proposed a crisis lexicon for sampling and filtering crisis-related messages in English during different emergency events. 
\citet{ChileEarthquakeT1} studied user and content based features to classify relevant tweets to an earthquake in Spanish (73.4\% F1-score) using Random Forest.
\citet{CrisisNlpR7} proposed a deep learning framework based on semi-supervised learning to classify relevant messages in English. 
They used two Twitter datasets, one of the Nepal earthquake (65.11\% F1) and another of the Queensland floods (93.54\% F1).
%
\citet{li2018comparison} proposed a feature-based adaptation framework, which considers pre-trained and crisis-specific word embeddings, as well as sentence embeddings and different supervised classifiers.
They evaluated two classification tasks in English 
reaching an average accuracy of 88.5\%. 
They noted that crisis-specific embeddings were more suitable for more specific crisis-related tasks ({\em informative} vs {\em non-informative}), while the pre-trained embeddings were more suitable for more general tasks ({\em relevant} vs {\em  non-relevant}).
\citet{alam2019} performed a comparative study among various algorithms used to classify crisis-related messages.
%
Their results show competitive results between Support Vector Machines, Random Forest and Convolutional Neural Networks.


In view of the difficulty of comparing results, models and techniques in this research area, \citet{Benchmarking} developed a standard dataset based on existing data and provided train/dev/test partitions.
In addition, the authors provided benchmark results on English messages for informative (binary) and humanitarian (multi-class) classification tasks using deep learning algorithms.
\citet{HumAID2021} created a large-scale dataset of English-language tweets, which is composed of 19 disaster events that occurred between 2016 and 2019. The authors report the results of the classification of humanitarian information using classical and deep learning algorithms. They achieved an average weighted F1 of 78.1\% with the RoBERTa model.

\subsection{Transfer Learning}
Knowledge transfer or transfer learning involves two main concepts, {\em domain} and {\em task} \cite{pan2009survey}. 
A domain consists of a feature space and a marginal probability distribution, while a task consists of a label space and an objective predictive function.
It aims to use knowledge from a source domain and learning task to 
improve the learning of the 
predictive function in the target domain, where source and target domains or tasks differ.
The authors categorize it as inductive, transductive, and unsupervised, based on the settings between source and target. 
In transductive transfer learning, tasks are the same, but domains are different. It can have two variants \cite{pan2009survey,ruder2019neural}: 
\begin{itemize}
    \item Cross-lingual adaptation: domains have different feature spaces (e.g., documents written in two different languages). 
    \item Domain adaptation: domains have different marginal probability distributions (e.g., documents discuss different topics).
\end{itemize}

\noindent{\bf Cross-Lingual Adaptation.}
This variant has been studied in different Twitter message classification tasks in a multilingual manner, such as election analysis, emotion recognition, crisis detection, among others. 
In crisis message classification, 
\citet{espol2019} compared traditional and deep learning models using sparse representations and word embeddings to classify earthquake-related conversations in English and Spanish. 
%
The best cross-lingual results were using a Long Short Term Memory model including multilingual stacked embeddings, reporting a macro F1-score of 85.88\% from Spanish to English and 77.49\% from English to Spanish. 
\citet{DBLP_Lorini} evaluated pre-trained 
language-agnostic and language-aligned word embeddings with Convolutional Neural Networks for classification of flood-related messages in German, English, Spanish, and French. 
They compared a mono-lingual classifier, a cross-lingual classifier with {\em cold start} (using no training data in the target language), and a cross-lingual classifier with {\em warm start} (using 300 labeled instances in the target language).
They showed that both types of word embeddings could be used to classify a new language for which few or no labels are available. 
However, including a small set of data from the same target language improved the cross-lingual classification.  
Another approach was proposed by \citet{Khare2018}, who considered messages in English, Italian and Spanish from 30 crisis events of different types. 
They proposed a statistical-semantic crisis representation, extracting semantic relationships from BabelNet and DBpedia knowledge bases. They achieved a cross-lingual classification F1-score of 59.9\% on average.
In addition, \citet{Sarioglu2020} trained cross-lingual models based on contextual embeddings such as BERT, RoBERTa
and XLM-R, to detect urgency messages in low-resource languages, specifically in Sinhala and Odia.

\smallskip 
\noindent{\bf Domain Adaptation.}
Regarding domain adaptation using Twitter data, \citet{agrawal2018deep} 
analyzed cyberbullying detection across Twitter, Wikipedia and Formspring. 
They reported poor performance when transferring knowledge from Twitter to the other two datasets reaching around 10\% F1-score. 
Similarly, \citet{ARANGO2020101584} evaluated the cross-dataset classification of hateful tweets, showing that the models do not generalize well over different Twitter datasets obtaining an F1-score of just 21\% for the positive class.
In the crisis classification problem, \citet{imran2016} validated the impact of adding training examples from a different domain than the target. 
%
They classified messages related to earthquakes and floods published in several languages. 
%
%
Their experiments showed that in scenarios where there is not enough data, increasing training examples with tweets from other languages can be useful if both are very similar (e.g., Italian and Spanish). 
In the case of domain adaptation, they concluded that using tweets from a different domain did not appear to improve performance. 
\citet{Li2020} proposed a domain adaptation with reconstruction to classify disaster tweets. This adaptation based on semi-supervised learning aims to reduce the shift between source and target data distributions. For this, the classifier is trained with labeled source data, together with unlabeled target data.

\smallskip
\noindent{\em{\bf Differentiation from prior work.}}
Our work differs in that we perform a systematical study of transfer learning for crisis message classification for scenarios in which little to no data is available. 
We focus on the case of how to leverage labeled data from high-resource languages to low-resource languages, as well as from well-known crisis domains to new domains.
We do not focus on the classification algorithms, but on the data and experimental methodology.
Moreover, we investigate which document representations and models work best for each particular target, putting together the most comprehensive dataset to date for this goal.

\section{Proposed Approach}\label{sec:proposal}

We propose an experimental analysis of transfer learning approaches across several crisis domains and different languages. 
In detail, we address the following specific objectives:  
1) determine if it is possible to transfer knowledge from one language to another,
2) determine if it is possible to transfer knowledge from one crisis domain to another domain, and
3) explore how to achieve (1) and (2) effectively.

We focus on the task of classifying microblog messages into the binary categories {\em related} and {\em not related} to crisis events\footnote{We consider {\em an event as a unique occurrence in time and place} \cite{enlighten128478}.}, within several transfer learning scenarios.
%
We adopt the definition used by \citet{review_article} that considers a message as {\em related} to a crisis when it {\em refers implicitly or explicitly to the specific disaster event for which it was retrieved}.
However, we generalize this definition to consider messages as {\em related} if they meet the above criteria for {\em any disaster}, not only a particular specified event.
In addition, we define a \textit{domain} as a {\em type of disaster (or hazard)} such as earthquakes or floods.

\begin{figure}[t]
  \centering
  \includegraphics[width=\columnwidth]{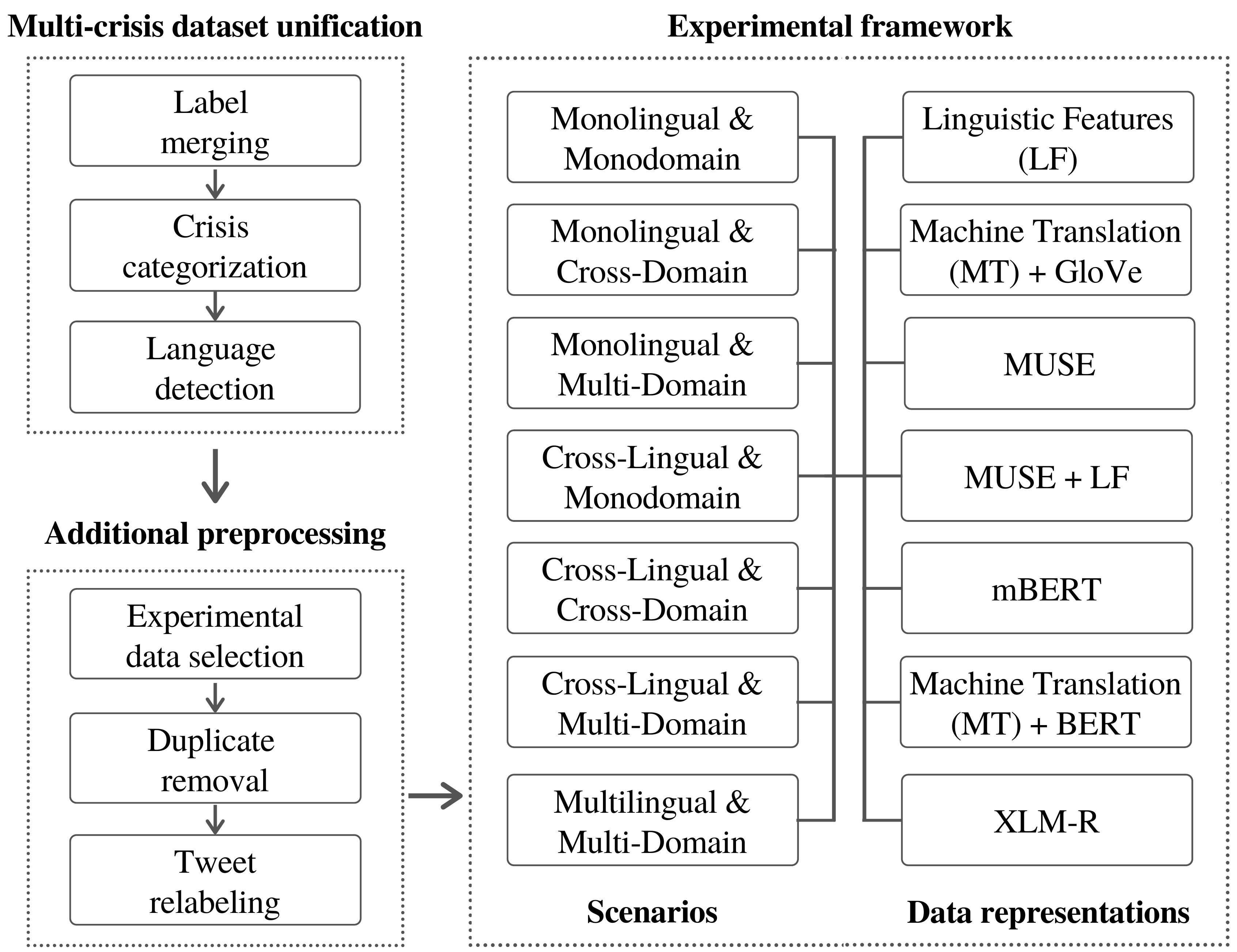}
  \caption{Overview of our workflow. This consists of the unification of the multi-crisis dataset, additional data preprocessing for our experiments, and experimental framework, including the transfer learning scenarios and data representations.} 
  
  \label{fig:diagram_pipeline}
\end{figure}

Figure \ref{fig:diagram_pipeline} provides an overview of the main steps followed in this work, both for dataset construction and experimentation. We explain these steps more in detail next.
%

\subsection{Transfer Learning Scenarios}\label{sec:transfer-scenarios}
We study 7 different classification scenarios that span across several transfer learning configurations.
These scenarios aim to measure the impact on learning--of both domains and languages--for the classification of new crisis events.
We include as baseline the scenario where there is no domain or language transfer learning.
%
To avoid overfitting, we split the training and testing data at event level, ensuring that events that are used for training are not used for testing.
%
%
%
More in detail:
%
\begin{itemize}
    \item {\bf Monolingual \& Monodomain.} 
    This is our baseline. 
    In this scenario, we train a classification model with messages from events of a specific crisis domain (e.g., earthquakes) all in the same language (e.g., English).
    We evaluate the model on messages from the same original domain and language, but from new events. 
    %
    
    \item {\bf Monolingual \& Cross-Domain.} 
    In this scenario, we only perform domain transfer learning within the same language. 
    The objective is to evaluate the effect of training with data from past
    events of a specific domain and language (e.g., earthquakes in English) to then classify messages from a different domain (e.g., explosions in English).
    This scenario evaluates the cold-start problem at domain level.
    This is, when a new type of disaster occurs for which there is no domain-specific labeled data.
    

    \item {\bf Monolingual \& Multi-Domain.} 
    In this scenario we increase the amount of training data for a specific (low-resource) domain by using additional, same language, labeled data from other domains. 
    %
    %
    For instance, we train a model using English earthquake, flood and explosion data. 
    We then evaluate this model on English data from a new flood event.  
    
    \item {\bf Cross-Lingual \& Monodomain.} 
    In this scenario, we perform cross-lingual adaptation within the same domain.
    The objective is to evaluate the effect of training a classification model with data from the target domain (e.g., floods) in a language (e.g., English) that is different from the target language (e.g., Italian).
    This simulates the real-world scenario in which there is only labeled data from a high-resource language for one domain and a new event occurs in the same domain, but in a different language.
    %
    %
    %
    \item {\bf Cross-Lingual \& Cross-Domain.} 
    In this scenario we perform cross-lingual and domain adaptation.
    The objective is to evaluate the effect of training a classification model with data from a set of crisis domains (e.g., earthquakes and floods) in one language (e.g., English). 
    Then, use this model to classify data of a different domain (e.g., explosions) in another language (e.g., Spanish).
    This simulates the case in which there is a need to classify messages of a new crisis domain in a language for which there is no labeled data.
    
    \item {\bf Cross-Lingual \& Multi-Domain.} 
    In this scenario we perform cross-lingual adaptation and increase training data using other domains.
    The objective is to evaluate the effect of training a model using data from multiple domains in one language (e.g., floods, earthquakes, and hurricanes in English), to then classify a new event in another language (e.g., flood in Italian).
    %
    %
    This simulates the case in which there is a need to classify a well known type of event that occurs in a new language for which there is no labeled data.
    \item {\bf Multilingual \& Multi-Domain.} 
    In this scenario we perform cross-lingual and domain data enrichment.
    This case is similar to the {\em Cross-lingual \& Cross-domain} scenario with the addition of data from the target domain and language.
    %
    %
    For example, we train a model with English and Italian data about floods, earthquakes, and hurricanes, to then classify flood-related messages in Italian.
    
\end{itemize}

\subsection{Data Representations}\label{sec:models}
We focus on message representations and state-of-the-art models that allow us to convey multilingual data in a single feature space.
Our representations correspond to 7 approaches described below.
%

\begin{itemize}
\item {\bf Linguistic Features (LF).} This representation models each message as a set of linguistic features. 
%
We consider 48 features\footnote{The full description of the features can be found in our repository.} represented in numerical and binary form.
Some of these features have been previously used for classification of crisis messages \cite{Graf2018, Khare2018}.
These features describe traditional message characteristics such as the number of {\em characters}, {\em words}, {\em links}, {\em mentions}, {\em hashtags}, {\em question marks}, among others. 
Furthermore, we consider attributes that are specific to each language,
including {\em sentiment polarity}, 
{\em Named Entities} 
and {\em Part-of-Speech (POS)} \cite{polyglotner, polyglotpos}.
%
We also include binary features such as {\em has mention} and {\em has location}.

\item {\bf Machine Translation (MT) + GloVe.} 
This representation is based on modeling messages using English as a pivot language. Hence it implies translating all messages that are not written in the pivot language into English\footnote{We use the Google Translate API: \url{https://cloud.google.com/translate/}}.
We choose English as our starting point for two reasons: 1) the high-quality of pre-trained embeddings in English, and 2) the trade-off of translating other languages to English, which is predominant in our dataset, as opposed to the other way around.
Text in the pivot language is then tokenized and vectorized using the pre-trained GloVe\footnote{\url{https://nlp.stanford.edu/projects/glove/}} model with 100-dimensions, trained on tweets \cite{glove_we}.

\item {\bf MUSE.} This representation uses MUSE\footnote{\url{https://github.com/facebookresearch/MUSE}}. MUSE consists of multilingual language-aligned word embeddings of 300-dimensions, based on fastText embeddings trained on Wikipedia \cite{muse_we}. 
%

\item {\bf MUSE + Linguistic Features (MUSE+LF).} This representation is a combination of MUSE and LF features. We use this to evaluate if the combination of semantic and statistical features improves model performance. 

\item {\bf mBERT.} This representation models each message using the BERT-Base Multilingual Cased model\footnote{\url{https://huggingface.co/bert-base-multilingual-cased}} trained with the top-104 languages on Wikipedia \cite{devlin_Mbert}. 

\item {\bf Machine Translation (MT) + BERT.} This representation is similar to (MT) + GloVe, but instead of GloVe, we use BERT\footnote{\url{https://huggingface.co/bert-base-uncased}} which has been trained for English using Wikipedia and BookCorpus \cite{devlin_Mbert}. 

\item {\bf XLM-R.} This representation uses XLM-RoBERTa\footnote{\url{https://huggingface.co/xlm-roberta-base}}, a multilingual version of RoBERTa trained on 100 languages of CommonCrawl \cite{XLM-RoBERTa}. 

\end{itemize}

For Glove and MUSE, we combine single word embedding using mean aggregation, and for words Out-Of-Vocabulary, we create vectors with zero. On the other hand, when using mBERT, BERT and XLM-R (all 768-dimension models, with no fine-tuning), we combine the contextualized word embeddings using mean aggregation, ignoring the padding of zeros.

\section{Unified Dataset Construction} \label{sec:data_unification} 
We created a new dataset that is the result of collecting 7 crisis datasets from the literature (see Table \ref{tab:datasets} for details). 
These datasets met the criteria of having labeled data and being publicly available.
The consolidated dataset, which we refer to as {\bf Multi-Crisis Dataset}\footnote{This dataset can be found at \url{https://github.com/cinthiasanchez/Crisis-Classification/tree/main/data/unified}}, contains Twitter messages from several crisis domains and events in different languages.
%
%
To unify labels and to achieve an enriched, consistent dataset, we performed a process that consisted of several data curation steps.
These mainly consisted of {\em label merging}, {\em crisis categorization}, and {\em language detection}. 
Next, we provide relevant details of this process.\footnote{The in-depth procedure can be found in our repository.}

\begin{table}[t]
	\centering
	{
	\resizebox{\columnwidth}{!}{%
		\fontsize{8pt} {6pt} \selectfont
		\begin{tabular*}{\columnwidth}{@{}p{0.50\columnwidth} p{0.15\columnwidth} r r @{}}
			\toprule
			\textbf{Dataset}  & \textbf{Lang.} & \textbf{Tweets} & \textbf{Events} \\
			\midrule
			\begin{tabular}[c]{@{}l@{}} CrisisLexT6 \\ \cite{CrisisLex6} \end{tabular}  & English & 60,082 & 6 \\ \midrule
			
			\begin{tabular}[c]{@{}l@{}} CrisisLexT26 \\ \cite{CrisisLex26} \end{tabular} & \begin{tabular}[c]{@{}l@{}} Multiple \end{tabular}  & 27,933 & 26 \\  \midrule 
			
			\begin{tabular}[c]{@{}l@{}} CrisisNLP\_R1 \\ \cite{CrisisNlpR1} \end{tabular} & \begin{tabular}[c]{@{}l@{}} English, \\Spanish,\\ French\end{tabular}  & 49,596 & 14 \\ \midrule 
			
			\begin{tabular}[c]{@{}l@{}} Ecuador-Earthquake \\ \cite{espol2019} \end{tabular} & \begin{tabular}[c]{@{}l@{}} English,\\ Spanish \end{tabular} & 8,360  & 1 \\   \midrule 
			
			\begin{tabular}[c]{@{}l@{}} SoSItalyT4 \\ \cite{SoSItalyT4} \end{tabular} & Italian & 5,642 &  4\\   \midrule
			
			\begin{tabular}[c]{@{}l@{}} ChileEarthquakeT1 \\ \cite{ChileEarthquakeT1} \end{tabular} &  Spanish & 2,187  & 1 \\  \midrule 
			
			\begin{tabular}[c]{@{}l@{}} CrisisMMD \\ \cite{crisismmd} \end{tabular} & English &  11,400 &  7\\  \midrule
			
			\begin{tabular}[c]{@{}l@{}} \textbf{Multi-Crisis Dataset}\\ \textbf{(Ours)} \end{tabular} & \textbf{Multiple} & \textbf{164,625} & \textbf{53}\\ 
			\bottomrule
	\end{tabular*}}
	}
\caption{Datasets used to create our Unified Multi-Crisis Dataset.}
\label{tab:datasets}
\end{table}

\medskip
\noindent{\bf Label Merging. }
We based our dataset unification process on prior work in the area of crisis informatics that has addressed the label merging problem. In particular, in order to merge the 7 different datasets used in our study we followed the systematic label mapping approach used by \citet{Khare2018, Benchmarking, hsarmiento2021}.
This methodology maintains data consistency by unifying labels into a more general class according to their semantic meaning \cite{Benchmarking}.
Specifically, the datasets used in our study had different types of labels, both binary and categorical at diverse granularities. We therefore merged these labels at a higher generalization level, which in this case was their relatedness to crisis events (i.e., {\em related} and {\em not-related} messages). These new labels correspond with the binary task required for our current work, and are similar to those already used in the {\em CrisisLexT6, ChileEarthquakeT1} and {\em Ecuador-Earthquake} datasets.

For instance, examples of original labels that we mapped to the {\em not-related} category in our dataset were: {\em ``not related''}, {\em ``not relevant''}, {\em ``not related or irrelevant''}, {\em ``off-topic''}, {\em ``not informative''}, {\em ``not applicable''}, and {\em ``not physical landslide''}. 
On the other hand, examples of labels mapped to the \emph{related} category were: {\em ``related''}, {\em ``relevant''}, {\em ``damage''}, {\em ``no damage''}, {\em ``on-topic''}, {\em ``informative''}, {\em ``related and informative''}, {\em ``related but not informative''}, among others. 
Furthermore, we manually relabeled messages from the CrisisMMD and CrisisLexT26 datasets originally labeled as {\em ``not informative''} and {\em ``Not applicable''} to accurately match our current task definition. The detailed mappings and relabeled data can be found at \url{https://github.com/cinthiasanchez/Crisis-Classification/blob/main/data/experimental%20subset.md}.

\medskip
\noindent{\bf Crisis Categorization. }
We annotated each message according to the crisis dimensions of the event that it belongs to. We used a similar pipeline to \citet{CrisisLex26}.
We categorized crises by hazard types (such as {\em ``earthquake''} or {\em ``explosion''}), hazard categories ({\em ``natural''} or {\em ``human-induced''}), sub-categories (e.g.,  {\em ``geophysical''}, {\em ``hydrological''}, {\em ``accidental''}, etc.), temporal development ({\em ``instantaneous''} or {\em ``progressive''}), and geographic spread ({\em ``focalized''} or {\em ``diffused''}).
We decided to include this information 
because it will allow the study of communication patterns along the different dimensions. For example, to study the similarities and differences in communication during natural events versus human induced events.
Furthermore, we aggregated information about the crisis, such as country and year. 

\medskip
\noindent{\bf Unified Dataset Description. }
Our dataset is mainly composed of messages in English (83.67\%), followed by Spanish (7.30\%) and Italian (4.25\%). With regard to the domain, it is mainly composed of earthquake (25.47\%), flood (19.39\%) and hurricane (11.89\%) domains. Table \ref{tab:unified_dataset_messages} shows more information on the number of messages per language and domain.

\begin{table}[t]
    \begin{subtable}{.5\linewidth}
      \centering
        
        \fontsize{7.5pt} {5pt} \selectfont
\begin{tabular}{@{}lrr@{}}
\toprule
\multicolumn{1}{c}{\textbf{Lang.}} & \multicolumn{1}{c}{\textbf{Count}} & \multicolumn{1}{c}{\textbf{\%}} \\ \midrule
English & 137,743 & 83.67 \\ \midrule
Spanish & 12,025 & 7.30 \\ \midrule
Italian & 7,002 & 4.25 \\ \midrule
French & 1,144 & 0.70 \\ \midrule
Portuguese & 771 & 0.47 \\ \midrule
Tagalog & 502 & 0.31 \\ \midrule
Russian & 238 & 0.15 \\ \midrule
German & 124 & 0.08 \\ \midrule
Indonesian & 111 & 0.07 \\ \midrule
Dutch & 101 & 0.06 \\ \midrule
Others & 4,864 & 2.96 \\ \midrule
\textbf{Total} & \textbf{164,625} & \textbf{100.00} \\ \bottomrule
\end{tabular}
\caption{Messages by language.}

    \end{subtable}%
    \begin{subtable}{.5\linewidth}
      \centering

        \fontsize{7.5pt} {5pt} \selectfont
\begin{tabular}{@{}lrr@{}}
\toprule
\multicolumn{1}{c}{\textbf{Domain}} & \multicolumn{1}{c}{\textbf{Count}} & \multicolumn{1}{c}{\textbf{\%}} \\ \midrule
Earthquake & 41,931 & 25.47 \\ \midrule
Flood & 31,923 & 19.39 \\ \midrule
Hurricane & 19,578 & 11.89 \\ \midrule
Typhoon & 13,674 & 8.31 \\ \midrule
Explosion & 12,004 & 7.29 \\ \midrule
Bombings & 11,012 & 6.69 \\ \midrule
Tornado & 9,992 & 6.07 \\ \midrule
Landslide & 4,492 & 2.73 \\ \midrule
Wildfires & 3,533 & 2.15 \\ \midrule
Viral disease & 3,512 & 2.13 \\ \midrule
Others & 12,974 & 7.88 \\ \midrule
\textbf{Total} & \textbf{164,625} & \textbf{100.00} \\ \bottomrule
\end{tabular}
\caption{Messages by domain.}

    \end{subtable} 
    \caption{Number and percentage of messages in the Unified Multi-Crisis Dataset detailed by (a) language and (b) crisis domain separately.}
    \label{tab:unified_dataset_messages}
\end{table}

\section{Experimental Setup} \label{sec:setup_train_test}
%
In this section we explain our experimental setup for evaluating the proposed transfer learning scenarios and data representations (see Section \ref{sec:models}).
%
For our experiments we used a portion of the Multi-Crisis Dataset, selecting messages from the top-3 most represented languages in the dataset (English, Spanish and Italian).
%
%
%
In addition, we discarded messages corresponding to events that 1) contained very little data, or 2) were from hazard domains not available for more than one language.
%
%
This resulted in a dataset with 67,001 tweets from various regions in 3 languages and from 3 hazard domains: earthquakes (46.9\%), floods (38.4\%) and explosions (14.7\%).
Additionally, our experimental dataset contains 80.0\% of English messages, 11.3\% in Spanish, and 8.7\% in Italian.
Regarding the label distribution, 36.0\% were categorized as {\em not-related} to crisis while 64.0\% were labeled as {\em related}.

For evaluation, each event was distributed into a training or test set.
To provide more representative examples for generalization, training sets prioritized events with the highest and most balanced number of instances in our dataset.
We did not use English as a target language for cross-language evaluation because it is a high-resource language.
The detailed training/testing partitions for each evaluation scenario can be found at \url{https://github.com/cinthiasanchez/Crisis-Classification/blob/main/data/experimental%20subset.md}.

Tables \ref{tab:training_set_scenario_monolingual} and \ref{tab:training_set_scenario_cross-multi_lingual} show the number of instances that were used for model training across scenarios, grouped by target language and domain. 
This data displayed an important imbalance between the positive {(\em related)} and negative {\em (not related)} classes, where the positive was approximately 27\% more represented than the negative class.
To handle this we applied random subsampling and oversampling of the positive and negative classes, respectively.

We trained all models using a Random Forest classifier from the Scikit-Learn implementation\footnote{\url{https://scikit-learn.org/stable/modules/generated/sklearn.ensemble.RandomForestClassifier.html}} 
by considering the following default parameters: $\mathit{n\_estimators=100}$ that represents the number of trees in the forest, $\mathit{gini}$ criterion, and $\mathit{no limit}$ for the maximum tree depth, so nodes expand until all leaves are pure or until all leaves contain less than two samples.
We chose Random Forest because previous works have shown competitive results for tweet classification tasks and practical applications \cite{alam2019, li2018comparison, ChileEarthquakeT1, ARANGO2020101584,HumAID2021}.
Nevertheless, the scope of our current work is not on comparing classification algorithms, but on classifier initialization approaches for low-resource languages, which could help train more robust classifiers.

\begin{table}[t]
	\centering
	{
	\resizebox{0.90\columnwidth}{!}{%
		\fontsize{9pt} {6pt} \selectfont
		\begin{tabular}{@{}llrrr@{}}
        \toprule
        \multicolumn{1}{c}{\textbf{Lang.}} & \multicolumn{1}{c}{\textbf{Domain}} & \textbf{\begin{tabular}[c]{@{}r@{}}Monolingual \&\\ Monodomain\end{tabular}} & \textbf{\begin{tabular}[c]{@{}r@{}}Monolingual \&\\ Cross-Domain\end{tabular}} & \textbf{\begin{tabular}[c]{@{}r@{}}Monolingual \&\\ Multi-Domain\end{tabular}} \\ \midrule
        \multirow{3}{*}{English} & Earthquake & 11,214 & 35,720 & 46,238 \\ \cmidrule(l){2-5} 
         & Explosion & - & 51,662 & - \\ \cmidrule(l){2-5} 
         & Flood & 10,346 & 34,418 & 46,916 \\ \midrule
        \multirow{2}{*}{Spanish} & Earthquake & 2,822 & 150 & - \\ \cmidrule(l){2-5} 
         & Explosion & - & 4,182 & - \\ \midrule
        \multirow{2}{*}{Italian} & Earthquake & 1,520 & 414 & - \\ \cmidrule(l){2-5} 
         & Flood & - & 2,114 & - \\ \bottomrule
	\end{tabular}}
	}
\caption{Number of training instances by target (language and domain) used in the monolingual scenarios, considering the balanced sets. The symbol ``-'' means that no experiments were performed.}
\label{tab:training_set_scenario_monolingual}
\end{table}

\begin{table}[t]
	\centering
	{
	\resizebox{\columnwidth}{!}{%
		\fontsize{9pt} {6pt} \selectfont
		\addtolength{\tabcolsep}{-3.0pt} 
		\begin{tabular}{@{}llrrrr@{}}
			\toprule
            \multicolumn{1}{c}{\textbf{Lang.}} & \multicolumn{1}{c}{\textbf{Domain}} & \textbf{\begin{tabular}[c]{@{}r@{}}Cross-Lingual \&\\ Monodomain\end{tabular}} & \textbf{\begin{tabular}[c]{@{}r@{}}Cross-Lingual \&\\ Cross-Domain\end{tabular}} & \textbf{\begin{tabular}[c]{@{}r@{}}Cross-Lingual \&\\ Multi-Domain\end{tabular}} & \textbf{\begin{tabular}[c]{@{}r@{}}Multilingual \&\\ Multi-Domain\end{tabular}} \\ \midrule
            \multirow{2}{*}{Spanish} & Earthquake & 18,324 & 35,720 & 56,346 & 67,260 \\ \cmidrule(l){2-6} 
             & Explosion & 9,282 & 51,662 & 63,460 & 69,768 \\ \midrule
            \multirow{2}{*}{Italian} & Earthquake & 20,426 & 35,720 & 63,122 & 67,520 \\ \cmidrule(l){2-6} 
             & Flood & 26,674 & 34,418 & 63,244 & 67,686 \\ \bottomrule
	\end{tabular}}
	}
\caption{Number of training instances by target (language and domain) used in the cross-lingual and multilingual scenarios, considering the balanced sets.}
\label{tab:training_set_scenario_cross-multi_lingual}
\end{table}

Table \ref{tab:test_size_target} shows the number of instances related to the events that remained for model testing.
As in our training data, testing data also showed an important class imbalance.
Due to the smaller amount of messages in our negative class for testing, oversampling to improve class balance was not a feasible solution.
To deal with this situation we opted to augment the negative class by including a random sample of negative instances from the complete testing set (i.e., not only from the test event for that scenario).
As negative instances are by definition messages not related to any crises, this does not affect evaluation.
In the cases where this type of augmentation was not possible due to language restrictions for the scenario, we augmented with translated negative instances from English\footnote{Translated using Google Translate.}.

\begin{table}[t]
\centering
\fontsize{7.5pt} {5pt} \selectfont
\begin{tabular}{@{}llrr@{}}
\toprule
\multicolumn{1}{c}{\textbf{Lang.}} & \multicolumn{1}{c}{\textbf{Domain}} & \multicolumn{1}{c}{\textbf{Related}} & \multicolumn{1}{c}{\textbf{Not related}} \\ \midrule
\multirow{3}{*}{English} & Earthquake & 8,611 & 1,225 \\ \cmidrule(l){2-4} 
 & Explosion & 4,415 & 4,641 \\ \cmidrule(l){2-4} 
 & Flood & 8,272 & 5,747 \\ \midrule
\multirow{2}{*}{Spanish} & Earthquake & 2,507 & 453 \\ \cmidrule(l){2-4} 
 & Explosion & 747 & 50 \\ \midrule
\multirow{2}{*}{Italian} & Earthquake & 698 & 198 \\ \cmidrule(l){2-4} 
 & Flood & 1,759 & 138 \\ \bottomrule
\end{tabular}
\caption{Original data available for testing for each target (before augmentation).}
\label{tab:test_size_target}
\end{table}

\section{Results} \label{sec:results_language}
We present a comprehensive summary of the results of our evaluation following our experimental setup.
There are many additional results based on the combinations of scenarios, domains and languages, which we cannot include due to space constraints.
However, the exhaustive evaluation results can be found in our online repository\footnote{\url{https://github.com/cinthiasanchez/Crisis-Classification/tree/main/results/balanced}}
, including an evaluation of XLM-T\footnote{\url{https://huggingface.co/cardiffnlp/twitter-xlm-roberta-base}} (model fine-tuned to multilingual Twitter data), which was outperformed by XLM-R.

\begin{figure}[h]
  \centering
  \includegraphics[width=\linewidth]{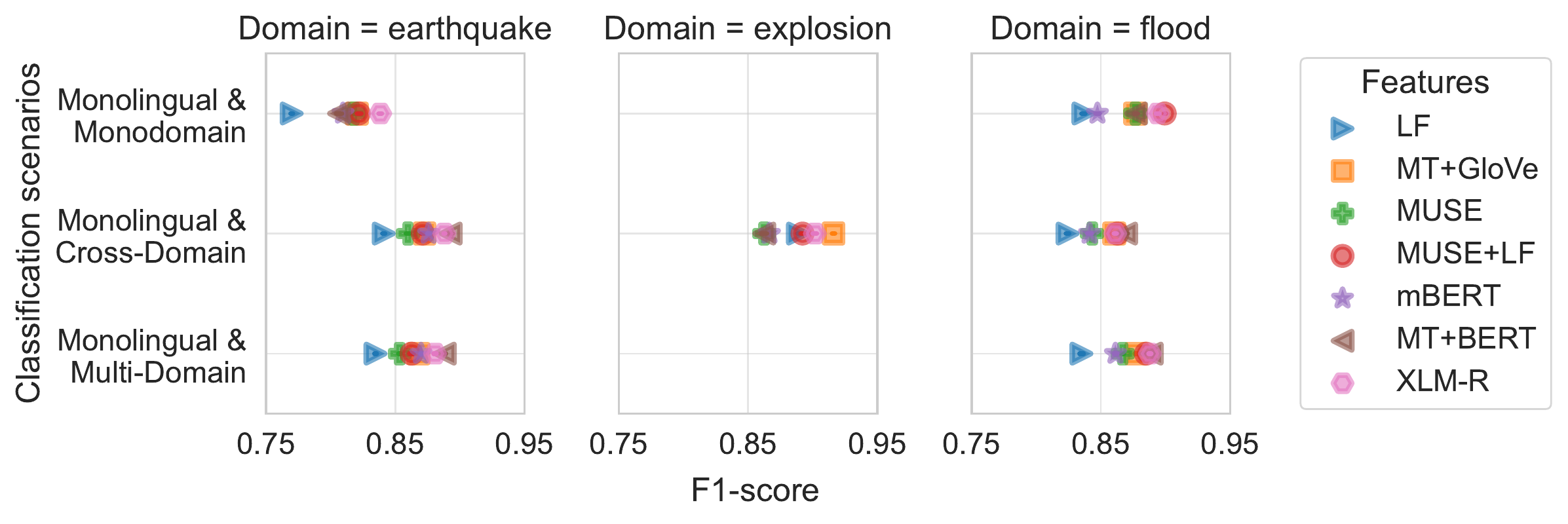}
  \caption{Results for English classification.
  } 
  \label{fig:english_results_plot}
\end{figure}

\begin{figure}[h]
  \centering
  \includegraphics[width=\linewidth]{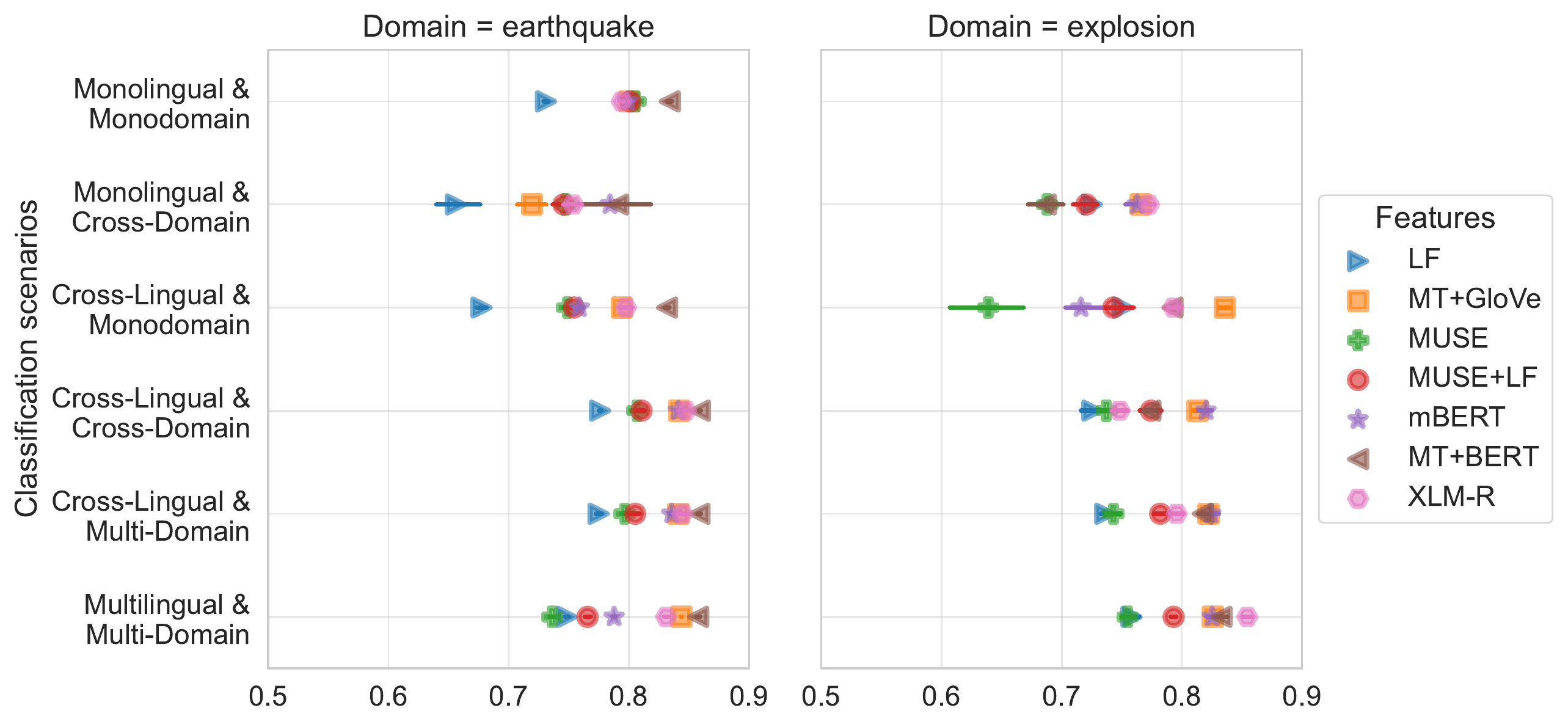}
  \caption{Results for Spanish classification.
  } 
  \label{fig:spanish_results_plot}
\end{figure}

\begin{figure}[h]
  \centering
  \includegraphics[width=\linewidth]{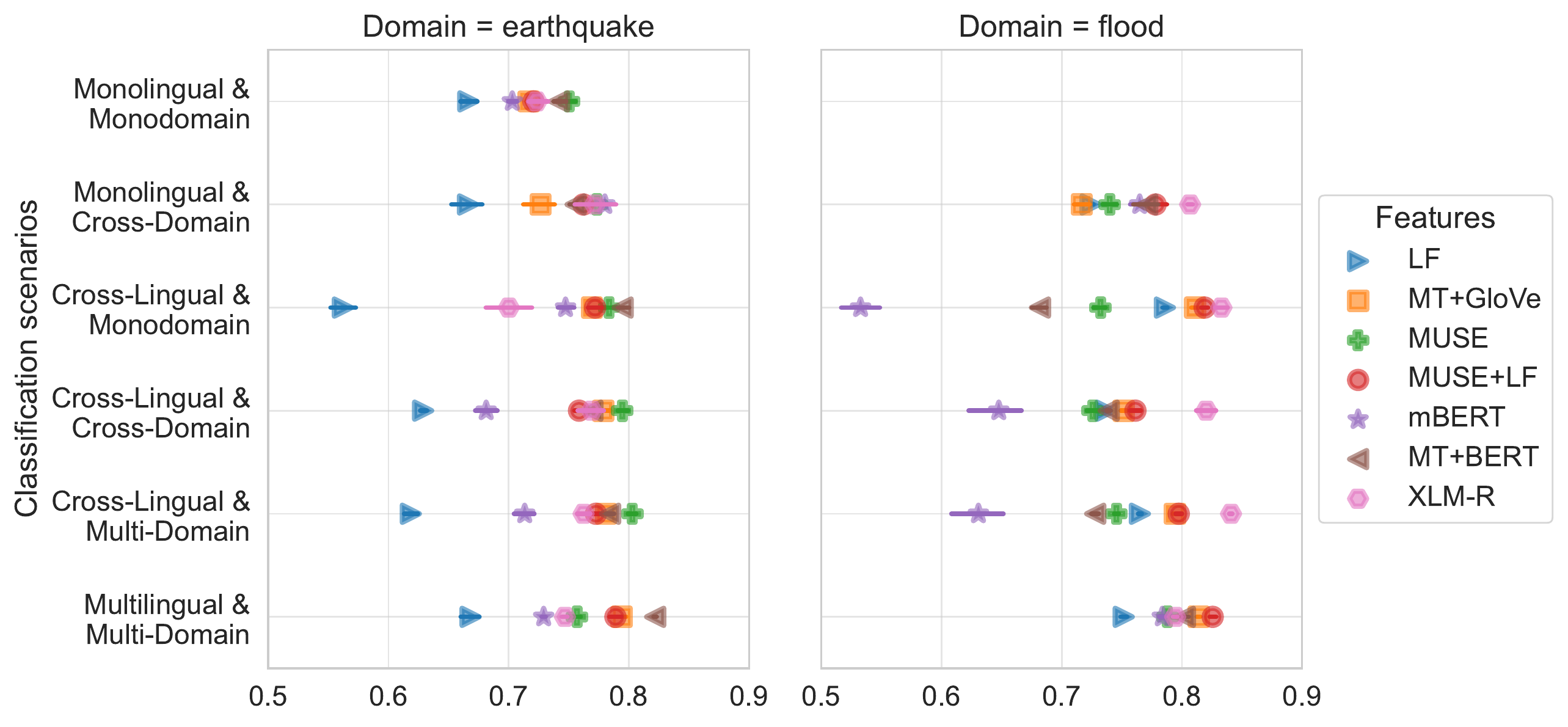}
  \caption{Results for Italian classification.
  } 
  \label{fig:italian_results_plot}
\end{figure}

We begin this section by describing the results obtained by target language and domain, running each model five times and averaging their values. 
These results are also shown in Figures \ref{fig:english_results_plot}, \ref{fig:spanish_results_plot} and \ref{fig:italian_results_plot} in Section \ref{sec:appendix} (Appendix).
Finally, we present an exhaustive analysis to determine the best performance for each classification scenario.

\subsection{English Classification}
We evaluated the classification of crisis related messages in English.
Specifically, monolingual scenarios for the same domain, cross-domain (domain adaptation) and multiple domains (data enrichment). We did not evaluate cross lingual scenarios with English as a target language as we consider it as a high-resource language in our setup.
However, we performed cross-lingual adaptation for low-resource languages using English as a source, detailed in the following sections.

Table \ref{tab:english_results} details the results obtained per scenario for each data representation and each of the three domains: earthquakes, explosions and floods, respectively.
The overall results show that models achieved their best performance by using XLM-R, followed by MT+GloVe and MT+BERT, while using LF, MUSE and mBERT did not work as well in this case.
However, mBERT's performance would probably improve if we not only use its representations (as with the other pre-trained models), but also use its capabilities to adapt the model to the crisis domain.

The best performing model for earthquake message classification (evaluation included messages from {\em Chile 2014, California, Pakistan} and {\em Ecuador} events) was 89\% F1-score in the cross-domain scenario.
In this scenario, we developed domain adaptation from explosion and flood domains.
In particular, this scenario obtained an average improvement of 5\% over the baseline scenario (monodomain).
For explosion message classification, our dataset only contained one event ({\em West Texas} event). 
Using this event as target we evaluated the only possible scenario, i.e., domain adaptation or cross-domain (training was done with earthquakes and floods), yielding an F1-score of 92\% with the best data representation (GloVe).
Recall that this scenario simulates the case in which we need to classify events from a new type of crisis event.

For flood message classification (evaluation included events of {\em Queensland, Pakistan} and {\em India}), the best scenario was the baseline with F1-score of 90\% using MUSE+LF. 
Although, cross-domain (87\% F1) showed good performance, similar to that of earthquake classification.
On the other hand, multi-domain (89\% F1) or domain enrichment achieved competitive results. 

Overall, we observe that domain adaptation appears to work well within a high-resource language.
This would allow us to use past knowledge to classify new and unexpected events in the same language.
Therefore, it indicates one could use a pre-trained classifier to detect new types of crises that emerge.
In addition, data augmentation (multi-domain) by including data from other domains can potentially improve model performance, or in the worst-case perform similar to the baseline.

\begin{table}[t]
\centering
\begin{subtable}[h]{0.45\textwidth}
\centering
\fontsize{7pt} {5pt} \selectfont
\addtolength{\tabcolsep}{-3pt} 

\begin{tabular}{@{}lccccccc@{}}
\toprule
\multicolumn{1}{c}{\textbf{Scenario}} & \textbf{LF} & \textbf{\begin{tabular}[c]{@{}c@{}}MT+\\ GloVe\end{tabular}} & \textbf{MUSE} & \textbf{\begin{tabular}[c]{@{}c@{}}MUSE+\\ LF\end{tabular}} & \textbf{mBERT} & \textbf{\begin{tabular}[c]{@{}c@{}}MT+\\ BERT\end{tabular}} & \textbf{XLM-R} \\ \midrule
\begin{tabular}[c]{@{}l@{}}Monolingual \& \\ Monodomain\end{tabular} & 0.77 & 0.82 & 0.82 & 0.82 & 0.81 & 0.81 & \textbf{0.84} \\ \midrule
\begin{tabular}[c]{@{}l@{}}Monolingual \& \\ Cross-Domain\end{tabular} & 0.84 & 0.87 & 0.86 & 0.87 & 0.88 & \cellcolor[HTML]{D8D7D7}\textbf{0.89} & \cellcolor[HTML]{D8D7D7}\textbf{0.89} \\ \midrule
\begin{tabular}[c]{@{}l@{}}Monolingual \& \\ Multi-Domain\end{tabular} & 0.83 & 0.87 & 0.85 & 0.86 & 0.87 & \cellcolor[HTML]{D8D7D7}\textbf{0.89} & 0.88 \\ \bottomrule
\end{tabular}

\caption{F1-score (Earthquake)} \label{tab:english_earthquake}
\end{subtable}

\hfill
    
\begin{subtable}[h]{0.45\textwidth}
\centering
\fontsize{7pt} {5pt} \selectfont
\addtolength{\tabcolsep}{-3pt} 

\begin{tabular}{@{}lccccccc@{}}
\toprule
\multicolumn{1}{c}{\textbf{Scenario}} & \textbf{LF} & \textbf{\begin{tabular}[c]{@{}c@{}}MT+\\ GloVe\end{tabular}} & \textbf{MUSE} & \textbf{\begin{tabular}[c]{@{}c@{}}MUSE+\\ LF\end{tabular}} & \textbf{mBERT} & \textbf{\begin{tabular}[c]{@{}c@{}}MT+\\ BERT\end{tabular}} & \textbf{XLM-R} \\ \midrule
\begin{tabular}[c]{@{}l@{}}Monolingual \& \\ Cross-Domain\end{tabular} & 0.89 & \cellcolor[HTML]{D8D7D7}\textbf{0.92} & 0.86 & 0.89 & 0.87 & 0.86 & 0.90 \\ \bottomrule
\end{tabular}

\caption{F1-score (Explosion)}
\label{tab:english_explosion}
\end{subtable}

\hfill
    
\begin{subtable}[h]{0.45\textwidth}
\centering
\fontsize{7pt} {5pt} \selectfont
\addtolength{\tabcolsep}{-3pt} 

\begin{tabular}{@{}lccccccc@{}}
\toprule
\multicolumn{1}{c}{\textbf{Scenario}} & \textbf{LF} & \textbf{\begin{tabular}[c]{@{}c@{}}MT+\\ GloVe\end{tabular}} & \textbf{MUSE} & \textbf{\begin{tabular}[c]{@{}c@{}}MUSE+\\ LF\end{tabular}} & \textbf{mBERT} & \textbf{\begin{tabular}[c]{@{}c@{}}MT+\\ BERT\end{tabular}} & \textbf{XLM-R} \\ \midrule
\begin{tabular}[c]{@{}l@{}}Monolingual \& \\ Monodomain\end{tabular} & 0.84 & 0.88 & 0.88 & \cellcolor[HTML]{D8D7D7}\textbf{0.90} & 0.85 & 0.88 & 0.89 \\ \midrule
\begin{tabular}[c]{@{}l@{}}Monolingual \& \\ Cross-Domain\end{tabular} & 0.82 & 0.86 & 0.84 & 0.86 & 0.84 & \textbf{0.87} & 0.86 \\ \midrule
\begin{tabular}[c]{@{}l@{}}Monolingual \& \\ Multi-Domain\end{tabular} & 0.83 & 0.88 & 0.87 & 0.88 & 0.86 & \textbf{0.89} & \textbf{0.89} \\ \bottomrule
\end{tabular}

\caption{F1-score (Flood)}
\label{tab:english_flood}
\end{subtable}

     \caption{Comparison of the data representations' performance in (a) earthquake, (b) explosion, and (c) flood domains for English message classification. For each domain, the best data representation is highlighted in bold and the best scenario is in a gray cell.}
     \label{tab:english_results}
\end{table}

\begin{table}[t]
\centering
\begin{subtable}[h]{0.45\textwidth}
\centering
\fontsize{7pt} {5pt} \selectfont
\addtolength{\tabcolsep}{-3pt} 
\begin{tabular}{@{}lccccccc@{}}
\toprule
\multicolumn{1}{c}{\textbf{Scenario}} & \textbf{LF} & \textbf{\begin{tabular}[c]{@{}c@{}}MT+\\ GloVe\end{tabular}} & \textbf{MUSE} & \textbf{\begin{tabular}[c]{@{}c@{}}MUSE\\ +LF\end{tabular}} & \textbf{mBERT} & \textbf{\begin{tabular}[c]{@{}c@{}}MT\\ +BERT\end{tabular}} & \textbf{XLM-R} \\ \midrule
\begin{tabular}[c]{@{}l@{}}Monolingual \& \\ Monodomain\end{tabular} & 0.73 & 0.80 & 0.81 & 0.80 & 0.80 & \textbf{0.83} & 0.79 \\ \midrule
\begin{tabular}[c]{@{}l@{}}Monolingual   \& \\ Cross-Domain\end{tabular} & 0.66 & 0.72 & 0.75 & 0.75 & 0.78 & \textbf{0.79} & 0.75 \\ \midrule
\begin{tabular}[c]{@{}l@{}}Cross-Lingual   \& \\ Monodomain\end{tabular} & 0.68 & 0.79 & 0.75 & 0.75 & 0.76 & \textbf{0.83} & 0.80 \\ \midrule
\begin{tabular}[c]{@{}l@{}}Cross-Lingual \& \\ Cross-Domain\end{tabular} & 0.78 & 0.84 & 0.81 & 0.81 & 0.84 & \cellcolor[HTML]{D8D7D7}\textbf{0.86} & 0.85 \\ \midrule
\begin{tabular}[c]{@{}l@{}}Cross-Lingual   \& \\ Multi-Domain\end{tabular} & 0.77 & 0.84 & 0.80 & 0.81 & 0.84 & \cellcolor[HTML]{D8D7D7}\textbf{0.86} & 0.84 \\ \midrule
\begin{tabular}[c]{@{}l@{}}Multilingual   \& \\ Multi-Domain\end{tabular} & 0.75 & 0.84 & 0.74 & 0.77 & 0.79 & \cellcolor[HTML]{D8D7D7}\textbf{0.86} & 0.83 \\ \bottomrule
\end{tabular}
\caption{F1-score (Earthquake)} \label{tab:spanish_earthquake}
\end{subtable}

\hfill
    
\begin{subtable}[h]{0.45\textwidth}
\centering
\fontsize{7pt} {5pt} \selectfont
\addtolength{\tabcolsep}{-3pt} 
\begin{tabular}{@{}lccccccc@{}}
\toprule
\multicolumn{1}{c}{\textbf{Scenario}} & \textbf{LF} & \textbf{\begin{tabular}[c]{@{}c@{}}MT+\\ GloVe\end{tabular}} & \textbf{MUSE} & \textbf{\begin{tabular}[c]{@{}c@{}}MUSE\\ +LF\end{tabular}} & \textbf{mBERT} & \textbf{\begin{tabular}[c]{@{}c@{}}MT\\ +BERT\end{tabular}} & \textbf{XLM-R} \\ \midrule
\begin{tabular}[c]{@{}l@{}}Monolingual \& \\ Cross-Domain\end{tabular} & 0.73 & \textbf{0.77} & 0.69 & 0.72 & 0.76 & 0.69 & \textbf{0.77} \\ \midrule
\begin{tabular}[c]{@{}l@{}}Cross-Lingual   \& \\ Monodomain\end{tabular} & 0.75 & \textbf{0.84} & 0.64 & 0.74 & 0.72 & 0.79 & 0.79 \\ \midrule
\begin{tabular}[c]{@{}l@{}}Cross-Lingual \& \\ Cross-Domain\end{tabular} & 0.73 & 0.81 & 0.74 & 0.77 & \textbf{0.82} & 0.77 & 0.75 \\ \midrule
\begin{tabular}[c]{@{}l@{}}Cross-Lingual   \& \\ Multi-Domain\end{tabular} & 0.74 & \textbf{0.82} & 0.74 & 0.78 & \textbf{0.82} & \textbf{0.82} & 0.80 \\ \midrule
\begin{tabular}[c]{@{}l@{}}Multilingual   \& \\ Multi-Domain\end{tabular} & 0.76 & 0.83 & 0.75 & 0.79 & 0.83 & 0.83 & \cellcolor[HTML]{D8D7D7}\textbf{0.85} \\ \bottomrule
\end{tabular}
\caption{F1-score (Explosion)}
\label{tab:spanish_explosion}
\end{subtable}

     \caption{Comparison of the data representations' performance in (a) earthquake and (b) explosion domains for Spanish message classification. For each domain, the best data representation is highlighted in bold and the best scenario is in a gray cell.}
     \label{tab:spanish_results}
\end{table}

\subsection{Spanish Classification}
We consider this language as low-resource, since the amount of labeled data is significantly less than for English (see Table \ref{tab:unified_dataset_messages}).
Our experiments included monolingual, cross-lingual and multilingual scenarios.
Table \ref{tab:spanish_results} presents the results for earthquake and explosion domains, respectively, broken down by the data representations. Overall, MT+BERT and XLM-R obtained the best results, followed by MT+GloVe and mBERT, while LF and MUSE did not perform as well.

%
In the case of earthquake messages classification, the best performance scenarios with an 86\% F1-score (on events from {\em Ecuador, Guatemala} and {\em Costa Rica}) used MT+BERT features, improving over the ({\em Monolingual \& Monodomain}) baseline in average a 3\%.
We observed this improvement for the cross-lingual and multilingual scenarios that used some sort of cross-language adaptation from English.
Among these best scenarios, there is the {\em Cross-lingual \& Cross-domain} scenario, which simulates classifying messages of a new type of crisis domain in a new language. 
Hence, there does not appear to be additional improvement when including data from the target domain and from the target language.
In addition, the worst performance was for the {\em Monolingual \& Cross-domain} scenario, which simulates the case when we attempt to classify messages from a new domain in the same language.
This most likely occurs due to the small amount of training data in Spanish, which limits cross-domain learning within that language.

For the explosion message classification in Spanish, we only had one event's worth of data (an event in {\em Venezuela}).
Therefore, we were not able to evaluate the baseline scenario ({\em Monolingual \& Monodomain}).
We observed that the best performance scenario, with 85\% F1-score, was the {\em Multilingual \& Multi-domain} scenario using XLM-R as features, and data in English and Spanish as the source.
As with earthquakes, the worst performance scenario was training with another domain in the same language as the target (i.e., {\em Monolingual \& Cross-Domain}). The worst performance feature along all scenarios was MUSE. 
%
In general, our results indicate that baseline performance can be improved by augmenting a low-resource language with high-resource language data, including multiple domains.

\begin{table}[t]
\centering
\begin{subtable}[h]{0.45\textwidth}
\centering
\fontsize{7pt} {5pt} \selectfont
\addtolength{\tabcolsep}{-3pt} 
\begin{tabular}{@{}lccccccc@{}}
\toprule
\multicolumn{1}{c}{\textbf{Scenario}} & \textbf{LF} & \textbf{\begin{tabular}[c]{@{}c@{}}MT+\\ GloVe\end{tabular}} & \textbf{MUSE} & \textbf{\begin{tabular}[c]{@{}c@{}}MUSE+\\ LF\end{tabular}} & \textbf{mBERT} & \textbf{\begin{tabular}[c]{@{}c@{}}MT+\\ BERT\end{tabular}} & \textbf{XLM-R} \\ \midrule
\begin{tabular}[c]{@{}l@{}}Monolingual \& \\ Monodomain\end{tabular} & 0.67 & 0.72 & \textbf{0.75} & 0.72 & 0.70 & 0.74 & 0.72 \\ \midrule
\begin{tabular}[c]{@{}l@{}}Monolingual \& \\ Cross-Domain\end{tabular} & 0.67 & 0.73 & 0.77 & 0.76 & \textbf{0.78} & 0.76 & 0.77 \\ \midrule
\begin{tabular}[c]{@{}l@{}}Cross-Lingual \& \\ Monodomain\end{tabular} & 0.56 & 0.77 & 0.78 & 0.77 & 0.75 & \textbf{0.79} & 0.70 \\ \midrule
\begin{tabular}[c]{@{}l@{}}Cross-Lingual \&   \\ Cross-Domain\end{tabular} & 0.63 & 0.78 & \textbf{0.79} & 0.76 & 0.68 & 0.77 & 0.77 \\ \midrule
\begin{tabular}[c]{@{}l@{}}Cross-Lingual \&\\ Multi-Domain\end{tabular} & 0.62 & 0.78 & \textbf{0.80} & 0.77 & 0.71 & 0.78 & 0.76 \\ \midrule
\begin{tabular}[c]{@{}l@{}}Multilingual \& \\ Multi-Domain\end{tabular} & 0.67 & 0.79 & 0.76 & 0.79 & 0.73 & \cellcolor[HTML]{D8D7D7}\textbf{0.82} & 0.75 \\ \bottomrule
\end{tabular}

\caption{F1-score (Earthquake)} \label{tab:italian_earthquake}
\end{subtable}

\hfill
    
\begin{subtable}[h]{0.45\textwidth}
\centering
\fontsize{7pt} {5pt} \selectfont
\addtolength{\tabcolsep}{-3pt} 
\begin{tabular}{@{}lccccccc@{}}
\toprule
\multicolumn{1}{c}{\textbf{Scenario}} & \textbf{LF} & \textbf{\begin{tabular}[c]{@{}c@{}}MT+\\ GloVe\end{tabular}} & \textbf{MUSE} & \textbf{\begin{tabular}[c]{@{}c@{}}MUSE+\\ LF\end{tabular}} & \textbf{mBERT} & \textbf{\begin{tabular}[c]{@{}c@{}}MT+\\ BERT\end{tabular}} & \textbf{XLM-R} \\ \midrule
\begin{tabular}[c]{@{}l@{}}Monolingual \& \\ Cross-Domain\end{tabular} & 0.72 & 0.72 & 0.74 & 0.78 & 0.76 & 0.77 & \textbf{0.81} \\ \midrule
\begin{tabular}[c]{@{}l@{}}Cross-Lingual \& \\ Monodomain\end{tabular} & 0.79 & 0.81 & 0.73 & 0.82 & 0.53 & 0.68 & \textbf{0.83} \\ \midrule
\begin{tabular}[c]{@{}l@{}}Cross-Lingual \&   \\ Cross-Domain\end{tabular} & 0.74 & 0.75 & 0.73 & 0.76 & 0.65 & 0.74 & \textbf{0.82} \\ \midrule
\begin{tabular}[c]{@{}l@{}}Cross-Lingual \& \\ Multi-Domain\end{tabular} & 0.77 & 0.79 & 0.75 & 0.80 & 0.63 & 0.73 & \cellcolor[HTML]{D8D7D7}\textbf{0.84} \\ \midrule
\begin{tabular}[c]{@{}l@{}}Multilingual \& \\ Multi-Domain\end{tabular} & 0.75 & 0.81 & 0.79 & \textbf{0.83} & 0.78 & 0.80 & 0.79 \\ \bottomrule
\end{tabular}
\caption{F1-score (Flood)}
\label{tab:italian_flood}
\end{subtable}

     \caption{Comparison of the data representations' performance in (a) earthquake and (b) flood domains for Italian message classification. For each domain, the best data representation is highlighted in bold and the best scenario is in a gray cell. }
     \label{tab:italian_results}
\end{table}

\subsection{Italian Classification}
For Italian, we performed a similar evaluation to Spanish.
Table \ref{tab:italian_results} presents the classification results for earthquake and flood domains, respectively. 
We observed that the MT+BERT representation performed better in the classification of earthquake messages, while XLM-R performed better with the flood domain. 
The other representations showed behavior that varies according to the target domain. 
For example, LF features obtained the worst performance classifying earthquakes, and MUSE features obtained the best performance in most scenarios. 
However, such behavior was different when classifying floods. 

For earthquakes the best performance (on {\em L'Aquila} event) was achieved in the {\em Multilingual \& Multi-domain} scenario (82\% F1) using MT+BERT and for floods (on {\em Sardinia} and {\em Genova} events) it was in the {\em Cross-lingual \& Multi-domain} scenario (84\% F1) using XLM-R.
In general, the cross-lingual adaptation and augmentation using English (a high-resource language) improved performance. Also, adding multiple domains from English and the target language increased this improvement. 

\subsection{Results by Classification Scenario}
We present the aggregated results grouped by scenario and detailed by features as Table \ref{tab:general_results} shows.
We exclude the baseline scenario since not all targets were evaluated in that scenario due to the scarcity of data.
The best performance was achieved in the {\em Multilingual \& Multi-domain} scenario, i.e., when we trained our model with all available data in English and the target language from multiple domains, followed by the {\em Cross-lingual \& Multi-domain} scenario.
On the contrary, the lowest performance was obtained in the {\em Monolingual \& Cross-domain} scenario.
As for the data representations, we observed that the {\em baseline feature (LF)} presents the lowest score in all scenarios, while {\em MT+BERT} achieved the highest score. However, XLM-R and MT+GloVe performed better in most scenarios.
Overall, we found that increased training data with multiple domains in English language contributed positively to the construction of multilingual and multi-domain models. 
Given the constraints posed by the diversity of data sources and the disparities in the size of the data, this improved our results. However, with better quality data we may observe larger effects on the improvements.

\begin{table}
\centering
\fontsize{7pt} {5pt} \selectfont
\addtolength{\tabcolsep}{-3pt} 
\begin{tabular}{@{}lccccccc@{}}
\toprule
\multicolumn{1}{c}{\textbf{Scenario}} & \textbf{LF} & \textbf{\begin{tabular}[c]{@{}c@{}}MT+\\ GloVe\end{tabular}} & \textbf{MUSE} & \textbf{\begin{tabular}[c]{@{}c@{}}MUSE\\ +LF\end{tabular}} & \textbf{mBERT} & \textbf{\begin{tabular}[c]{@{}c@{}}MT\\ +BERT\end{tabular}} & \textbf{XLM-R} \\ \midrule
\begin{tabular}[c]{@{}l@{}}Monolingual \& \\ Cross-Domain\end{tabular} & 0.69 & 0.73 & 0.74 & 0.75 & 0.77 & 0.75 & \textbf{0.78} \\ \midrule
\begin{tabular}[c]{@{}l@{}}Cross-Lingual \& \\ Monodomain\end{tabular} & 0.69 & \textbf{0.80} & 0.73 & 0.77 & 0.69 & 0.78 & 0.78 \\ \midrule
\begin{tabular}[c]{@{}l@{}}Cross-Lingual \& \\ Cross-Domain\end{tabular} & 0.72 & \textbf{0.80} & 0.77 & 0.78 & 0.75 & 0.79 & \textbf{0.80} \\ \midrule
\begin{tabular}[c]{@{}l@{}}Cross-Lingual \& \\ Multi-Domain\end{tabular} & 0.72 & \textbf{0.81} & 0.77 & 0.79 & 0.75 & 0.80 & \textbf{0.81} \\ \midrule
\begin{tabular}[c]{@{}l@{}}Multilingual \& \\ Multi-Domain\end{tabular} & 0.73 & 0.82 & 0.76 & 0.79 & 0.78 & \cellcolor[HTML]{D8D7D7}\textbf{0.83} & 0.81 \\ \bottomrule
\end{tabular}
\caption{Average F1-score by classification scenario and feature for the low-resource languages Spanish and Italian, including their different crisis domains. The test examples are the same in all scenarios. The best result per scenario is highlighted in bold and the best overall result is in a gray cell.}
\label{tab:general_results}
\end{table}

\section{Discussion}\label{sec:discussion}
Our findings indicate that \emph{Multilingual \& Multi-domain} adaptation is an effective way to improve the classification of low-resource languages.
When no labeled data is available for a target language, a good option is to perform cross-lingual adaptation from a high-resource language using all available domain data.
Most importantly, we show that it is possible to classify messages that correspond to a new, previously unseen, crisis even when they are in an unknown language.
This can be very useful in identifying unexpected emerging crisis situations for early response. 

We propose and evaluate data augmentation (or enrichment) mechanisms that encompass a suite of techniques that enhance the size and quality of training datasets such that better models can be built using them. 
Data augmentation relies on creating a larger input data size to reduce the shift between the source and target distributions.
However, incorporating new data on training sets could decrease the accuracy of the classification models because of the lack of the ability to generalize adequately. 
In this sense, rather than adding noise when we included new content, our results indicated that data augmentation from multiple domains and languages could provide an opportunity in scenarios when data is insufficient for specific types of events.

A comparison of cross-domain results for Spanish and Italian reveals that transferring knowledge from one or more domains to another is useful from high-resource languages, such as English.
Nevertheless, it does not help if we use as a source low-resource languages (e.g., training with Spanish explosion to classify earthquakes in the same language).

Regarding the most effective ways to represent data for knowledge transfer, translating the content into English and using BERT or GloVe models provides the most accurate results for Spanish messages in most scenarios.
However, for Italian, XLM-R and MUSE+LF provide the best results.
This can be due to variability in the quality of machine translation, for example.
In practice, translating messages to English may not be cost-effective. 
However, this analysis allows us to understand the available alternatives.
In terms of F1-score, the observed difference between MT+GloVe, MT+BERT and XLM-R models is not statistically significant at a 95\% confidence level.
In addition, translating the content into English and then using the BERT model (trained in English corpus), obtained statistically superior results to using the mBERT model (trained on multilingual corpus).

Aligned embeddings, such as MUSE, are a promising less expensive approach for low-resource language classification.
We also observe that the LF representation was not as competitive across scenarios.
However, when this feature is combined with MUSE, it improves results, specifically for Italian.
Regarding mBERT, we show that it provides competitive results for some classification scenarios, but this is not consistent.
For future work, we will apply fine-tuning to this model 
(as well as to XLM-R) to explore whether the performance increases.

For explosion in Spanish and flood in Italian, we observed a decrease in performance for the {\em Cross-lingual \& Monodomain} scenario, using the MUSE and mBERT compared to LF.
This could be due to: 1) the dependence of specific words to the crisis domain and 2) the similarity of the representations of those words in both languages (English and the target language). The latter could be explained by the general corpus that was used to train MUSE and mBERT (Wikipedia).

\subsection{Analysis of Predictions}

\begin{figure}[t]
  \centering
  \includegraphics[width=\linewidth]{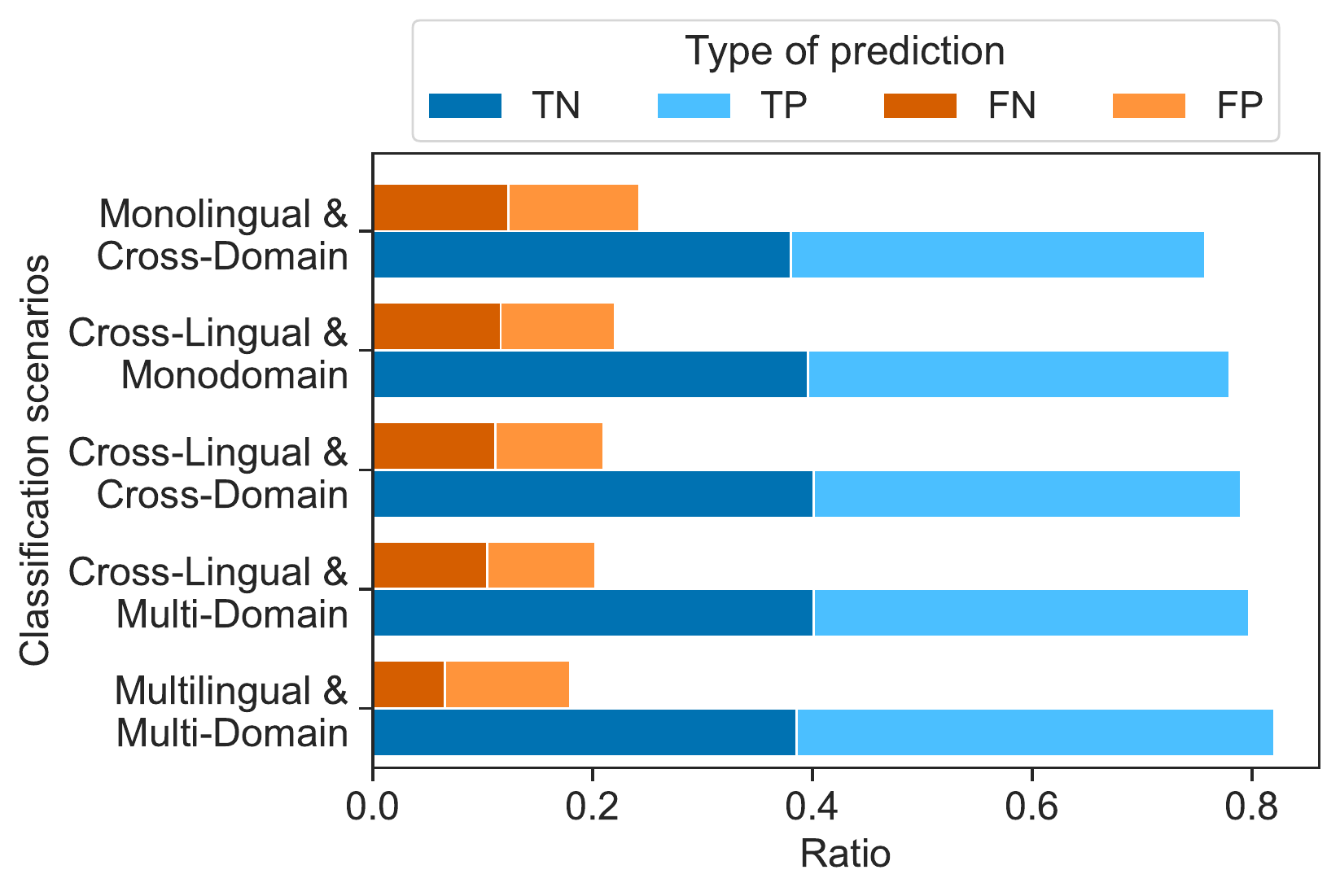}
  \caption{Predictions on Spanish and Italian test sets using the MT+BERT model. The test examples are the same in all scenarios. TP: True Positive, TN: True Negative, FP: False Positive, FN: False Negative.}
  \label{fig:prediction_results}
\end{figure}

We consider the analysis of the predictions to focus on the errors and suggest future improvements. 
Figure \ref{fig:prediction_results} shows the rate of correct and incorrect predictions for the Spanish and Italian languages using the MT+BERT model.  
We chose this model because it achieved the best overall performance for these languages (see Table \ref{tab:general_results}). 
We observe that the last scenario ({\em Multilingual \& Multi-domain)} presents the lowest error rate (FN + FP), where the rate of false negatives is lower than that of false positives.
These rates vary depending on the feature used. For example, with the MUSE+LF classifier, this scenario has a higher false positive-rate.

We performed a manual analysis of some classification errors. 
This analysis can provide insights into what might be the errors we need to address for future research.
Table \ref{tab:error_exam} shows some misclassified examples for all three languages.
Examples \textbf{(A)} and \textbf{(B)} are false negatives mainly belonging to the original label: \emph{related - but not informative} (CrisisLexT26), \emph{on-topic} (CrisisLexT6) and \emph{related} (Ecuador-Earthquake).
This reveals that false negatives are sometimes expressions of personal opinions, prayers, sympathy and emotional support.
%
%
Examples \textbf{(C)} and \textbf{(D)} show false positives mainly belonging to:  \emph{off-topic} (CrisisLexT6), \emph{not related} (CrisisLexT26) and \emph{not related or irrelevant} (CrisisNLP\_R1). 
In the first example, the message contains explicit information about an earthquake in Japan, but since it had been collected for the Ecuadorian earthquake, it was considered as {\em not related}, which could be considered as a mislabeled example for our task definition.
Likewise, \textbf{(D)} is related to a shooting event, but not to L'Aquila earthquake. 
In both cases, the classification model correctly marked them as being {\em related} to 
crises.
However, these messages were (incorrectly) labeled in the dataset as {\em not related}, because in their original datasets they were not related to the crisis that was being collected at the time.
%

We recognize within the limitations of our work, the unification of datasets from different sources. We have done our best to achieve coherent labels, but there may be inconsistencies.
%
%
On the other hand, automatic translation tools, such as google translate, 
may not be fully accurate in all scenarios.
There may be better translation tools but we did not have access to use them because of the high cost or difficulty with the API.
In terms of task, we evaluated our approach only on the binary task by relatedness to crises. We expect that our methodology allows our results to generalize well to other tasks, but we did not evaluate it in this work.

\begin{table}[t]
	\centering
	{
		\fontsize{9.0pt} {6.0pt} \selectfont
		\begin{tabular*}{\columnwidth}{@{}p{0.68\columnwidth}p{0.15\columnwidth} c@{}}
			\toprule
			\textbf{Tweet text} & \textbf{Crisis} & \textbf{Lang.} \\
			\midrule
			\textbf{(A)} \textit{Idk why these things are happening but I know god is with us !!!} & West Texas explosion & EN \\ \midrule
			
			\textbf{(B)} \textit{Svegliato dal terremoto, non sarà un bel giorno per molti.	
			\newline
			\textbf{Translation:} Awakened by the earthquake, it will not be a good day for many.	
			} & L'Aquila earthquake & IT\\  \midrule
			
			\textbf{(C)} \textit{Cifra de muertos en Japón aumenta a 41 tras terremoto: Equipos apuran el rescate https://t.co/uQJVU3ijDf
			\newline
			\textbf{Translation:} Death toll in Japan rises to 41 after earthquake: Teams rush to rescue https://t.co/uQJVU3ijDf 
			} & Ecuador earthquake & ES\\ \midrule
			
			\textbf{(D)} \textit{15 Germania: sparatoria in tribunale, almeno due morti e diversi feriti: Il conflitto a fuoco nell'aula di Lan.. http://tinyurl.com/cdb73r
			\newline
			\textbf{Translation:} 15 Germany: shooting in court, at least two dead and several injured: The firefight in the courtroom of Lan .. http://tinyurl.com/cdb73r	
			} & L'Aquila earthquake  & IT\\ 
			
			\bottomrule
	\end{tabular*}}
	
\caption{Examples of misclassified messages. 
\textbf{(A)} and  \textbf{(B)} represent false negatives and \textbf{(C)} and \textbf{(D)} false positives.}
\label{tab:error_exam}
\end{table}

\section{Conclusion}\label{sec:conclusion}
The main goal of this work was to assess the impact of transfer learning in the multilingual and multi-domain classification of messages relating to crises. 
In particular, data from a high-resource language such as English can contribute to the classification of messages from low-resource languages such as Spanish and Italian.
Furthermore, adding messages from the target language also helps in some cases.
Our findings indicate that there exist patterns in crisis communications that expand across crisis domains and languages.
As a result we can increase our ability to classify data in languages and domains for which we have little to no labeled data.
However, the most efficient data representations may vary depending on the target language.

For future work we want to explore in-depth different deep learning classifiers and techniques for cross-lingual and domain adaptation.
In addition, we would like to further improve our dataset by manually relabeling certain mislabeled messages that are effectively related to crises according to our task definition.
This will allow us to have a data collection that is oriented towards crises in general, rather than event oriented.
Moreover, we will explore this transfer learning approach in a more fine-grained task, such as categorizing actionable humanitarian information.

\subsection{Ethical Considerations}
We use existing social network data, clearly indicating the  source and respecting the policies on data sharing and anonymity. 
Our overarching goal is to expand crisis informatics applications to low-resource languages. 
However, there exist inequalities in access to the Internet, especially prevalent in low-income countries. These geographical and socio-demographic biases are present in the source of the data (social networks) and create important challenges for crisis informatics.

\subsection{Acknowledgments}
This work was funded by FONDECYT Project 1191604. Sánchez was also supported by SENESCYT-Ecuador Scholarship, and Sarmiento by ANID / Scholarship Program / DOCTORADO BECAS CHILE/2020 - 21201101. We also acknowledge the support of Millennium Institute for Foundational Research on Data, and the National Center for Artificial Intelligence CENIA FB210017, Basal ANID.
We also thank Aymé Arango, James Caverlee and Aidan Hogan for their useful comments.

{\footnotesize
\bibliography{references}}

\section{Appendix} \label{sec:appendix}

\end{document}